\pdfoutput=1

\documentclass[11pt]{article}

\usepackage[]{EMNLP2023}

\usepackage{times}
\usepackage{latexsym}
\usepackage{booktabs}
\usepackage{lipsum,graphicx,subcaption}
\usepackage{adjustbox}
\usepackage{mathtools}

\usepackage[T1]{fontenc}

\usepackage[utf8]{inputenc}

\usepackage{microtype}

\usepackage{inconsolata}
\usepackage{comment}
\usepackage{babel,blindtext}
\usepackage{cleveref}
\usepackage{geometry}
\usepackage{pdflscape}
\usepackage{rotating}

\title{Relevance-guided Neural Machine Translation}

\author{Isidora Chara Tourni \\
  Boston University\\
  \texttt{isidora@bu.edu} \\
  \And
  Derry Wijaya \\
  Boston University \\
  \texttt{wijaya@bu.edu}}

\begin{document}
\maketitle
\begin{abstract}
With the advent of the Transformer architecture, Neural Machine Translation (NMT) results have shown great improvement lately. 
However, results in low-resource conditions still lag behind in both bilingual and multilingual setups, due to the limited amount of available monolingual and/or parallel data; hence, the need for methods addressing data scarcity in an efficient, and explainable way, is eminent. We propose an explainability-based training approach for NMT, applied in Unsupervised and Supervised model training, for translation of three languages of varying resources, French, Gujarati, Kazakh, to and from English. Our results show our method can be promising, particularly when training in low-resource conditions, outperforming simple training baselines;  though the improvement is marginal, it sets the ground for further exploration of the approach and the parameters, and its extension to other languages. 
\end{abstract}

\section{Introduction}
Unsupervised Neural Machine Translation (UNMT) has seen remarkable progress in recent years, with a very large number of methods proposed aiming to NMT when parallel data are few or non-existent for certain language pairs \citep{artetxe2017unsupervised, lample2018phrase, conneau2017word, wang2021advances,lample2019cross,song2019mass,liu2020multilingual,
marchisio2020does,kim2020and,lample2017unsupervised,artetxe2019effective,garcia2020multilingual,su2019unsupervised, nguyen2022refining}.
Training techniques such as Back-Translation \citep{sennrich2015improving} and Auto-Encoding have been widely studied, in order to efficiently train NMT models under those data scarcity conditions to obtain high quality translation results. However, there is little work in enhancing Neural Machine Translation (NMT) models with utilizing explainability of the model in order to improve quality of the output. We propose a method, based on Layer-wise Relevance Propagation (LRP), which leverages the contribution of the input tokens to the output, to boost NMT performance. Our results show LRP may be beneficial during model training for NMT output improvement, particularly in low-resource conditions and for specific well defined model setups.

\section{Related Work}
\subsection*{Layer-wise Relevance Propagation (LRP)}
LRP 
was introduced by \citet{bach2015pixel}, measuring the contribution of the input components, or the neurons of a network, to the next layer's output. Due to its nature, it is directly applicable to layer-wise architectures, and we extend its usage to the Transformer architecture, measuring the relevance of source and target sentences' tokens to the NMT output during training.

\subsection*{Explanations \& Explanation-guided training}

Several previous works outline and summarize the findings of explainability and interpetability- related research in NLP \citep{belinkov-etal-2020-interpretability,sun2021interpreting,tenney2020language,madsen2021post,danilevsky2020survey,qian2021xnlp}.
\citet{weber2022beyond} provide a systematic review of explainable AI methods employed in improving various properties of Machine Learning models, such as performance, convergence, robustness, reasoning, efficiency and equality. Of these, of particular interest, and the focus of our work, are those that along with measuring feature importance and distinguishing relevant from irrelevant features, are utilized to augment the intermediate learned features, and improve model performance or reasoning \citep{anders2022finding,sun2021explanation,zunino2021excitation,fukui2019attention,zhou2016learning,mitsuhara2019embedding,schiller2019relevance,zunino2021explainable}. 

In our paper, we modify the approach of
\citet{sun2021explanation}, that proposes a model-agnostic LRP-guided training method for few shot classification, to improve model generalization to new classes, extending it to a transformer-based masked language model for NMT. Every intermediate feature representation f\textsubscript{p} is weighted (multiplied) by its relevance R(f\textsubscript{p}), with respect to the feature processing output, normalized in [-1,1]. The model is then trained on a loss function taking into account both predictions, $p$ and $p_{lpr}$ and given by the following formula
\begin{equation}
{L' = \xi * L (y,p) + \lambda * L (y, p\textsubscript{lrp})}\end{equation}where ${\xi}$, ${\lambda}$
are positive scalars. In this way, the features more relevant to the prediction are emphasized, while the less relevant ones are downscaled.
Other recent works utilize LRP for improving model performance in the medical domain
\citep{sefcik2021improving}, mitigating the influence of language
bias for image captioning models \citep{sun2022explain}.

\section{Method \& Experiments}


\subsection{Model \& Translation Quality Evaluation}
In our experiments we use a 6-layers 8-heads encoder-decoder transformer-based model, XLM \citep{lample2019cross}, following the training configurations and hyperparameters suggested by the authors. We use Byte Pair Encoding \citep{sennrich-etal-2016-neural} to extract a 60k vocabulary, and have an embedding layer size of 1024, a dropout value and an attention layer dropout value of 0.1, and a sequence length of 256. We measure the quality of the Language Model (LM) with perplexity, and quality of the NMT output with BLEU \citep{papineni2002bleu}, both used as training stopping criteria, when there is no improvement over 10 epochs. All further parameter values are provided in the Appendix.
We first pre-train a Language Model in each language with the MLM objective, which is then used to initialize the encoder and decoder of the NMT model. We further train a NMT model, 
using backtranslation (BT) and denoising auto-encoding (AE) with the monolingual data used for LM pretraining for UNMT, the Machine Translation (MT) objective for the Supervised NMT model, and BT+MT for the joint Unsupervised and Supervised approach.

\subsection{Datasets}
The languages we work with are English, French, Gujarati, Kazakh, and we're translating in all English-centric directions, English--French (En--Fr), French--English (Fr--En), English--Gujarati (En--Gu), Gujarati--English (Gu--En), English--Kazakh (En--Kk), Kazakh--English (Kk--En). For English and French, we use 5 million News Crawl 2007-2008 monolingual sentences for each language, and 23 million WMT14 parallel sentences. 
For Gujarati, we have 1.4 million sentences and for Kazakh we have 9.5M monolingual sentences, collected for both languages from Wikipedia, WMT 2018, 2019 and Leipzig Corpora (2016)\footnote{https://wortschatz.unileipzig.de/en/download/}. 
As parallel data, we have 22k and 132k from the WMT 2019 News Translation Task\footnote{http://data.statmt.org/news-crawl/} for Gu--En and Kk--En.
As development and test sets, we use newstest2013 and newstest2014, respectively, for En--Fr and Fr--En, WMT19 for En--Gu and Gu--En and En--Kk and Kk--En.

\begin{table*}[!htbp]
\centering
\small
\begin{tabular}{lcccc|c|ccc|c}
\toprule
&&& \textbf{En--Fr} & &&  & \textbf{Fr--En} &  \\\hline
Parameters' Set Values &  & v1 & v2 & v3 &Regular& v1 & v2 & v3 &Regular \\\hline
\midrule
22k  &&&&&&&&& \\
\midrule
MT  &&29.9&24.67&30.24&31.12&30.12&26.79 &\textbf{30.52}&30.63 \\
BT+AE+MT &&30.01&28.62&\textbf{32.15}&34.54&29.94&29.87&\textbf{33.03}&34.02 \\
\midrule
\midrule
1m  & &&&&&&&\\\hline
MT & &\textbf{39.12}& 38.12&35.44 &41.25& \textbf{39.2}&38.43&36.04&41.33
\\
BT+AE+MT & & \textbf{37.58}& 37.22& 36.89&40.37& 37.66& \textbf{37.95}& 37.38&40.4
\\
\midrule
\midrule
2.5m  &&&&&&&&\\\hline
MT &  &\textbf{39.06}&36.02  &35.57&40.46&\textbf{39.11} &36.28&36.09&40.71
\\
BT+AE+MT &  &37.68 & 36.21& \textbf{37.74} &39.88&40.71&37.48 &36.66& \textbf{38.15}
\\
\midrule
\midrule
5m &&&&&&&&\\\hline
MT  &&\textbf{39.21} &37.55&39&41.52&\textbf{39.33} &38.01&39.2&41.18
\\
BT+AE+MT &&30.71&36.7&\textbf{37.48}&40.89&32.02 &37.03&\textbf{37.67}&40.8
\\
\midrule
\midrule
{\emph{Other methods}} &&&&&45.9 &  &&&-\\
\bottomrule
\end{tabular}
\caption{BLEU scores for Supervised, and Unsupervised + Supervised NMT Layerwise Relevance Propagation-guided experiments, for En--Fr, Fr--En. \emph{AE}, \emph{BT} and \emph{MT} stand for Auto-Encoding loss, Back Translation loss and Machine Translation loss, respectively. Test and validation sets are from newstest2013-14 for French. State of the art results (\textit{Other methods}) for En--Fr come from \url{http://www.deepl.com/press.html}, \url{http://nlpprogress.com/english/machine_translation.html}.}
\vspace{-2mm}
\label{table:bleu_results_lrp_train_fr}
\vspace{-0.5em}
\end{table*}

\subsection{Layer-wise Relevance Propagation (LRP)}

We follow the method proposed by \citet{voita2020analyzing} in calculating LRP in an encoder-decoder Transformer architecture. The Relevance Score is first propagated inversely through the decoder and then the encoder, up to the input layer of the architecture. The conservation principle only holds across all processed tokens, and the score is defined as relevance of the input neurons to the top-1 logit predicted by the Transformer model, and the sum of the input neurons' relevance is regarded as the token contribution.
The total source and target sentence contributions to the result are defined as the summation of the Relevance of tokens in the source sentence, ${x}$ and that of those in the target sentence, ${y}$, at generation step t.
\begin{equation}
R_t(source)  = \sum_{i}^{}{x_i} ,
\\\\  \hspace{3mm}
R_t(target) =  \sum_{j=1}^{t-1}{y_j}
\end{equation}
At every step t, Relevance of the source and target sentences follow the conservation principle, summing up to 1.
Moreover, for every target token past the currently generated one, Relevance Score is 0.



\begin{table*}[!htbp]
\small
\centering
\begin{tabular}{lcccc|c|ccc|c}
\toprule
&&& \textbf{en--gu} &&&& \textbf{gu--en} && \\\hline
\midrule
Parameters' Set Values &  & v1 & v2 & v3 &Regular& v1 & v2 & v3 &Regular \\\hline
22k &&&&&\\
\midrule
MT && 2 & 2.18 &\textbf{2.19}&1.04&0.69&0.7 &\textbf{0.73}&2.65 \\
BT+AE+MT &&0.76&\textbf{1.36}&0.89&1.16 &0.71&\textbf{1.06}&0.67&2.19\\\hline
\midrule
{\emph{Other methods}} &&&&&0.1&  &&&0.3\\
\bottomrule
\end{tabular}

\caption{BLEU scores for Supervised, and Unsupervised + Supervised NMT Layerwise Relevance Propagation-guided experiments, for En--Gu, Gu--En. \emph{AE}, \emph{BT} and \emph{MT} stand for Auto-Encoding loss, Back Translation loss and Machine Translation loss, respectively. Test and validation sets are from WMT19 for Gujarati. State of the art results (\textit{Other methods}) can be found in \url{https://github.com/google-research/bert/blob/master/multilingual.md}. }
\vspace{-2mm}
\label{table:bleu_results_lrp_train_gu}
\vspace{-0.5em}
\end{table*}

\begin{table*}[!htbp]
\small
\centering
\begin{tabular}{lcccc|c|ccc|c}
\toprule
&& & \textbf{en--kk} &&&& \textbf{kk--en} && \\\hline
Parameters' Set Values &  & v1 & v2 & v3 &Regular& v1 & v2 & v3 &Regular \\\hline
\midrule
22k  &  && &&&   &&\\
\midrule
MT  &&2.1&3&\textbf{3.2}&2.4&2.2&2.1&\textbf{2.7}&2.6 \\
BT+AE+MT &&1.6&\textbf{3}&2.3&2.8&2.1&\textbf{2.6}&2.&2.9\\
\midrule
\midrule
132k  &&&&&&&&\\
MT &&4.8&5.6&5.3&5.2&6.8&\textbf{8.5}&8.4&8\\
BT+AE+MT &&5.2&\textbf{6.8}&6.4&6.6&8.7&\textbf{9.4}&9.2&8.9\\
\midrule
\midrule
{\emph{Other methods}} &&&&&2.5\footnote{https://www.deepl.com/press.html} &  &&&7.4\\
\bottomrule
\end{tabular}

\caption{BLEU scores for Supervised, and Unsupervised + Supervised NMT Layerwise Relevance Propagation-guided experiments, for En--Kk, Kk--En. \emph{AE}, \emph{BT} and \emph{MT} stand for Auto-Encoding loss, Back Translation loss and Machine Translation loss, respectively. Test and validation sets are from WMT19. State of the art results (\textit{Other methods}) can be found in \url{https://github.com/google-research/bert/blob/master/multilingual.md}.}
\vspace{-2mm}
\label{table:bleu_results_lrp_train_kk}
\vspace{-0.5em}
\end{table*}

\subsection{LRP-weighted training}
Following \citet{sun2021explanation}, we attempt to utilize LRP contributions during training, and examine performance. In our case, the representation of every intermediate source or target token $x_{i}$, with Relevance Score $R_t(x_i)$, is reweighted by its score at each layer, and included in a new loss term, $L\textsubscript{ce} (y,x_{i})$. The formula takes the following form
\begin{equation}
L' = \xi * L (y,p(x_i)) + \lambda * L (y, p({R_t}(x_i))\end{equation},where $p(x_i)$ and $p({R_t}(x_i))$ are the model prediction and the explanation-guided prediction, respectively.
The loss then is the weighted sum of the previous and the new terms, each weighted by parameters $\xi$, $\lambda$ respectively, for which we experiment with three sets of values:
\begin{equation}
\xi, \lambda = \{v\textsubscript{1} = \{1,0.5\}, v\textsubscript{2} = \{0,1\}, v\textsubscript{3} = \{1,1\}\}.
\end{equation}
In the first layer we only weigh the word embedding of the token.
We hypothesize that in this way, the tokens with a higher contribution to the NMT result are enhanced, while the effect of the ones contributing less is reduced.


\section{Results \& Discussion}
In Tables \ref{table:bleu_results_lrp_train_fr}, 
\ref{table:bleu_results_lrp_train_gu},
\ref{table:bleu_results_lrp_train_kk}, we present our results for LRP-guided training, in certain low- and high-resource Semi-Supervised and Supervised experiments, for all languages and directions, providing the regular NMT model results as our baselines.

We see that for En--Fr and Fr--En NMT, in Table \ref{table:bleu_results_lrp_train_fr}, the method fails to outperform baseline NMT results in all cases. The model translation quality is usually on par with baselines, and state-of-the art results on high scale experiments, and small differences in BLEU scores in the range of 0.1-0.5 can be considered negligible. 
Among the three hypermarameter settings, choosing v1 for training seems to outperform the other two in the majority of experiments under a MT-only setting. 
Results could indicate unsuitability of the method in high-resource settings; the original method was after all proposed in a few-shot classification context, hence we also seek more promising results in low-resource NMT experiments, examined below.

A different model behavior is observed for all cases but one in English to Gujarati NMT, in Table \ref{table:bleu_results_lrp_train_gu} when either the MT-only or BT-AE-MT objectives are used in training. This is an interesting finding - we can hypothesize that LRP-guided training might be more useful when translating into highly complex morphological languages such as Gujarati; Results marginally outperform previous state-of-the-art approaches, however more research including other languages and potentially other parameter values is required to verify that observation.

Encouraging is also the case of LRP-guided training in En--Kk and Kk--En NMT. In a large number of settings, training with Relevance guidance improves NMT BLEU scores significantly compared to our regular models' results. More specifically, we see improvement in both low- (22k) and mid-resource (132k) experiments, with v2-parameterized models to outperform our baselines in the majority of cases. Also, all experiments we are able to perform better than the state of the art current result, hence relevance guidance shows potential again in NMT experiments where few parallel data are available.

\section{Conclusions}
We perform a series of Semi-Supervised and Supervised Neural Machine Translation experiments, using an explainability-based metric, namely Relevance-guided propagation, during training; we leverage the measure of influence of the input and intermediate layer outputs to the NMT result, in an attempt to improve NMT for three quite different languages, lying in both high- and low- resource data regimes. Our results, though showing marginal and very small improvements, indicate that Layerwise-relevance propagation shows potential in boosting NMT quality when training in small data scenarios. Further exploration of the method, different model hyperparameter setups, and expansion of our method to other languages is strongly recommended as a next step to identify the efficiency and robustness of the proposed method.
\section{Limitations}
Training a large Neural Machine Translation model from scratch is a hard task computationally, 
and employing LRP-guidance during training significantly raises training time, the amount and usage of required computational resources, and the complexity of the training process, calling for more efficient training solutions, in terms of memory distribution of the model and parallelization. These factors constitute the limitations of our approach, and allowed us to launch a small number of experiments, hence addressing those factors and expanding to more languages, in more efficient training and computational ways, is a strong requirement for further generalization of the method.

\section{Ethics Statement}
Several ethical concerns ought to be addressed when working with large language models regarding quality, toxicity and bias related to their training process and output \cite{10.1145/3442188.3445922,chowdhery2022palm,brown2020language},of which the authors of the paper are aware in their work.

\newpage
\bibliography{anthology,custom}
\bibliographystyle{acl_natbib}

\end{document}